\documentclass[runningheads]{llncs}
\usepackage{iftex} 

\usepackage{eccv}

\usepackage{eccvabbrv}

\usepackage{graphicx}
\usepackage{booktabs}
\usepackage{xcolor}

\ifPDFTeX
\usepackage[accsupp]{axessibility}  
\fi

\usepackage{hyperref}

\usepackage{orcidlink}

\begin{document}

\title{From Interpretability Methods\\to Interpretable Models}

\titlerunning{From Methods to Interpretable Models}




\author{
Julien Colin\inst{1,2},
Nuria Oliver\inst{1}\textsuperscript{*} \and
Thomas Serre\inst{2,3}\textsuperscript{*}
}

\authorrunning{J.~Colin et al.}

\institute{
ELLIS Alicante, Spain
\and
Carney Institute for Brain Science, Brown University, USA
\and
Department of Cognitive and Psychological Sciences,
Brown University, USA
}

\maketitle

\begingroup
\renewcommand{\thefootnote}{\fnsymbol{footnote}}
\footnotetext[1]{Equal senior contribution. E-mail: julien\_colin@brown.edu}
\endgroup

\begin{abstract}
More than a decade in, explainable AI (XAI) for computer vision has assembled a mature toolbox: attribution, feature visualization, concept-based, and circuit-based methods. Yet almost all of the field's effort has gone into building and comparing these \emph{methods}, and little into the question they were meant to answer---how interpretable are our \emph{models}, and are we making progress as they evolve? We argue for shifting the field's focus from methods to models, along two complementary lines. One is already within reach: existing tools let us characterize and compare what different models represent and compute. The other is harder, and largely neglected: whether a model can actually be understood by the humans who rely on it—the independent evaluators on whom trust and certification depend, not the experts confirming what they already expect. It can only be measured, not inferred. We review why the toolbox is mature enough to support both, survey the thin body of work comparing models, draw a parallel to systems neuroscience, and close with a model-centric XAI agenda.
\end{abstract}

\section{Introduction}
Artificial intelligence now drives decisions in high-stakes domains: medical image interpretation~\cite{foundationmedical,d2025foundationradio,baharoon2023dinoradio}, autonomous driving~\cite{gao2025fsurvey,jiang2025fsurvey}, and robotic control~\cite{kim2025openvla}. These systems are accurate enough to deploy, yet remain largely black boxes to the people who rely on them. Understanding them nonetheless matters: to catch failures before they cause harm~\cite{geirhos2020shortcut,damour2022underspecification}, to meet emerging legal expectations of explanation~\cite{kaminski2021right}, and to learn from models that make new scientific discoveries~\cite{jumper2021alphafold,fawzi2022alphatensor}. As models grow more capable and more widely deployed, the gap between what they do and what we understand widens~\cite{hewitt2025vocabulary}, making interpretability a pressing need~\cite{kazmierczak2025explainability}. These pressures are beginning to acquire regulatory teeth: interpretability is increasingly named in certification frameworks---most explicitly by EASA, which lists ``AI explainability'' as a trustworthiness building block for aviation~\cite{easa2024conceptpaper}, and by the EU AI Act, whose high-risk provisions require systems transparent enough for humans to interpret and oversee their outputs~\cite{eu2024aiact}---an expectation that, while not yet a hard requirement, rises in proportion to risk.

What interpretability actually means, though, has never been settled: the term has been used for many distinct properties~\cite{lipton2016mythos,doshivelez2017rigorous}, and what they share is that it is ultimately human-centric---an explanation counts only if a person can actually understand it~\cite{miller2017explanation}. In computer vision, one effort has been to make models interpretable by design~\cite{rudin2019stop}\footnote{We retain \emph{interpretable by design} as it is the term used in the literature. But since we take interpretable to refer to a model that can be understood by humans, these methods are, in our view, closer to explainable by design: they guarantee that an explanation is available, not that the model can be understood.}: building transparency into the architecture itself---prototype networks that classify by comparison to visualizable exemplars~\cite{chen2019protopnet}, concept bottleneck models that route predictions through human-named concepts~\cite{koh2020concept}, or self-explaining networks that emit an explanation as part of the forward pass~\cite{alvarezmelis2018robust}---trading architectural flexibility for transparency. A separate effort---explainable AI (XAI)~\cite{gilpin2018explaining}, the focus of this paper---instead treats a trained model as a black box and explains its predictions after the fact. The line between the two is blurrier than it looks: designing a model to be explainable does not, on its own, make it interpretable.

Over the past decade and a half, XAI has produced several complementary families of methods---attribution, feature visualization, concept-based, and, more recently, circuit-based. Much of the community's effort, including work using human evaluations to ask how well a method lets us understand a model~\cite{colin2022cannot,kim2021hive,sixt2022users}, went into making these methods better and comparing them against one another, rather than into using them to explain and compare models.

This paper takes a different position. We do not claim that XAI methods are perfect, nor that method development should stop---only that the toolbox is now rich enough to redirect much of the field's effort from \emph{how well a method explains a model} to \emph{how interpretable a model is, and how that changes as models evolve}. That shift has two sides: comparing what different models represent and compute, which existing tools already support, and measuring whether an independent evaluator can actually understand them---which expert analysis alone cannot establish. The moment is right: methods are mature enough to build on (Section~\ref{sec:matured}), the model landscape is consolidating around a few families (Section~\ref{sec:now}), and---as systems neuroscience showed, building decades of insight with far cruder instruments---progress follows from studying the object rather than waiting for perfect tools.  
Yet remarkably little work turns the toolbox on models themselves (Section~\ref{sec:sparse}), leaving the field's original question open; we close with an agenda for a model-centric XAI (Section~\ref{sec:agenda}).

\section{The XAI toolbox has matured}\label{sec:matured}
The case for studying models rather than methods rests on a premise: the methods must be good enough to build on. This section argues that they are. We review the toolbox through three established families of XAI methods in computer vision, together with a fourth, more recent one that targets a model's computation rather than its representations. These four form a deliberately selected core of complementary approaches, rather than an exhaustive taxonomy of XAI. Each answers a different interpretability question, and each has matured enough to serve as a reliable instrument for comparing models, even if none is a complete account of how a model reasons.

\paragraph{\textbf{Attribution methods.}} Given a model, an input, and its prediction, which pixels drove this decision? This is perhaps the most fundamental interpretability question in computer vision, and the first to be systematically addressed by XAI through attribution methods~\cite{simonyan2013deep}. Their goal is to identify the image regions most responsible for a model's prediction, and a wide range of approaches has been developed over the past decade to do so: gradient-based~\cite{simonyan2013deep,shrikumar2017learning,smilkov2017smoothgrad,sundararajan2017axiomatic}, activation-based~\cite{selvaraju2017gradcam,chattopadhay2018grad}, perturbation-based~\cite{simonyan2014vgg,ribeiro2016lime,petsiuk2018rise,fel2021sobol}. The same question extends naturally beyond the final output to internal activations, where attribution instead identifies which inputs drive a given intermediate computation~\cite{dhamdhere2019how,leino2018influence}. 
The question can also be turned around: instead of asking what drove a prediction, counterfactual explanations ask what minimal change to the input would produce a different one~\cite{wachter2017counterfactual,goyal2019counterfactual}. In vision, however, the desired change is difficult to define: minimizing it risks an adversarial perturbation~\cite{freiesleben2022intriguing,jeanneret2023adversarial}, while enforcing a meaningful change---increasingly through generative models~\cite{augustin2022diffusion,jeanneret2022diffusion}---risks sacrificing faithfulness to the model being explained~\cite{khan2024faithful}. As a result, counterfactual explanations are not yet established as a reliable general-purpose interpretability tool.

\vspace{1mm}
Attribution methods themselves have arguably received the most scrutiny of the four families, from builders and critics alike. Early sanity checks revealed that several popular methods behaved no differently from edge detectors and were insensitive to model or label randomization~\cite{adebayo2018sanity}. Subsequent work exposed additional shortcomings related to reliability~\cite{kindermans2019reliability}, robustness and manipulability~\cite{ghorbani2017interpretation,dombrowski2019manipulated}, evaluation methodology~\cite{yeh2019infidelity,hooker2018benchmark}, the very definition of faithfulness these methods are meant to satisfy~\cite{jacovi2020towards}, and, more recently, poor transfer from convolutional networks to transformer and multimodal architectures~\cite{chefer2021transformer,kazmierczak2025explainability}. This sustained criticism has progressively narrowed the field toward methods with stronger theoretical grounding, such as those satisfying axiomatic properties~\cite{sundararajan2017axiomatic,lundberg2017unified} or based on principled statistical foundations~\cite{fel2021sobol,bach2015pixel}. Consequently, recent advances have tended to refine this core set of methods rather than introduce fundamentally new approaches.

\vspace{1mm}
Although attribution methods remain imperfect and continue to improve, they have already proven useful for helping humans understand model behavior. In particular, users can leverage attribution maps to reliably predict a model's behavior on unseen inputs~\cite{colin2022cannot},
as well as to improve performance in visual search and human-AI teaming tasks~\cite{kim2021hive,nguyen2022visual}. Taken together, this evidence suggests that attribution methods are sufficiently mature to serve as reliable tools for comparing models, even if they are not perfect explanations of model reasoning.

\paragraph{\textbf{Feature visualization methods.}} While attribution methods are useful, they have an intrinsic limitation: they only explain \emph{where} the model pays attention, not \emph{what} it pays attention to there. Feature visualization methods offer a complementary way to address this gap~\cite{mahendran2015understanding,nguyen2016multifaceted,olah2017feature,nguyen2019understanding}, either by showing natural images that highly activate a target element, or by generating a synthetic image to maximize its activation. The latter approach offers more flexibility, as it goes beyond dataset examples, but this flexibility comes at a cost: without properly constraining the optimization process, it tends to produce either adversarial-looking patterns or stimuli that might be visually appealing but unfaithful to what is actually encoded~\cite{nguyen2016multifaceted,nguyen2016synthesizing,nguyen2017plug}. Recent work has largely addressed this weakness by constraining the optimization to a small set of parameters and imposing a human-derived prior directly on the input, allowing the approach to scale to today's leading vision models~\cite{fel2023unlocking}, although interpreting increasingly large foundation model can remain challenging. Both variants have since been evaluated for how well they let humans understand what a unit encodes, with encouraging results~\cite{borowski2021,zimmermann2021,zimmermann2023,colin2024choosing}. A complementary line forgoes synthesized images and instead describes what a unit encodes directly in natural language, in vision and language models alike~\cite{hernandez2022natural,oikarinen2023clipdissect,bills2023language}.

\paragraph{\textbf{Concept-based methods.}} Attribution and feature visualization can explain any target, but historically, that target was individual neurons in a given layer. This proved too narrow, or too entangled, a basis for many units~\cite{bau2017network,olah2017feature,fong2018net2vec}. TCAV~\cite{kim2018interpretability} proposed a different basis instead: directions derived from human-defined concepts. One major weakness of this method is that it limits the vocabulary of explanations to whichever concepts are tested, which may not match those that the model relies on. ACE~\cite{ghorbani2019towards} addressed this by automating concept discovery directly from the model's representations, an approach later extended by ICE~\cite{zhang2021invertible}, CRAFT~\cite{fel2023craft}, and sparse autoencoders (SAE)~\cite{gao2024scaling}. 
All of these automated methods can be viewed through the unifying lens of dictionary learning~\cite{mairal2014sparse}, with different instantiations making different assumptions about how a model organizes its concepts~\cite{fel2023holistic,hindupur2025dual}. Nevertheless, these methods are already useful: they support direct intervention on models~\cite{dhimoila2026iso}, clarify which concepts different tasks recruit~\cite{fel2026into}, expose the systematic blind spots of other models~\cite{bohacek2025blindspots}, and, more broadly, help humans understand models~\cite{fel2023craft}.

\paragraph{\textbf{Circuit-based methods.}} The newest family moves from \emph{what} a model represents to \emph{how} it computes---reverse-engineering the specific circuits, the connected sets of features and weights, that implement a given behavior~\cite{elhage2021mathematical}. It was pioneered on vision models by the Circuits program, which traced curve detectors and other mechanisms through InceptionV1~\cite{olah2020zoom,olah2020an,cammarata2020curve,cammarata2020thread}, and more work has followed since~\cite{hamblin2022pruning,rajaram2024automatic,kwon2025granular,zukowska2026seeing,thasarathan2026cross}, though the approach has received comparatively more attention in language models. There, sparse autoencoders decompose representations into interpretable features at scale~\cite{gao2024scaling,bricken2023monosemanticity,cunningham2023sparse}, and circuit analysis has isolated reusable, human-legible mechanisms: induction heads that copy patterns in context~\cite{olsson2022induction}, a circuit for indirect-object identification~\cite{wang2023ioi}, and localized, editable factual associations~\cite{meng2022rome}. The multimodal case sits in between: recent work frames the failures of vision-language models through the binding problem~\cite{campbell2024binding} and reads their symbol-like mechanisms directly from activations~\cite{assouel2025visualsymbolic}. In vision proper, the approach is still comparatively niche, but it completes the toolbox: where attribution, feature visualization, and concepts describe a model's \emph{representations}, circuit-based methods aim to explain its \emph{computation}.

\section{Why Now}\label{sec:now}
\paragraph{\textbf{Imperfect tools can support cumulative science.}}
The use of imperfect but informative tools is a recurring practice in science. We take neuroscience as an example. As early as the late 1950s, Lettvin and colleagues read bug- and dimming-detectors directly off the frog's optic nerve~\cite{lettvin1959frog}; Hubel and Wiesel mapped orientation-selective receptive fields in cat visual cortex with single electrodes and slits of light cast from a slide projector~\cite{hubel1959receptive,hubel1995eye}, a strikingly simple setup by modern standards; and, further up the ventral stream, Gross and colleagues found inferotemporal neurons so selective they were discovered almost by accident, including a cell tuned to a hand and, famously, one that fired when the experimenters waved an actual toilet brush in front of the animal~\cite{gross1972inferotemporal}. Later work uncovered highly selective neurons in the medial temporal lobe~\cite{quiroga2005invariant,quiroga2008sparse,quiroga2013brain} in addition to distributed representations for faces and objects in ventral temporal cortex~\cite{haxby2001distributed}. The instruments sharpened in step, from hand-crafted stimuli to reverse correlation for mapping receptive fields~\cite{ringach2004reverse}, and on to causal tools such as microstimulation~\cite{salzman1990microstimulation} and lesion-based double dissociations~\cite{goodale1992separate} that intervene on a circuit rather than only observing it. Throughout, progress came from using whatever instrument was available and refining models as new findings exposed their limits.

This precedent comes with a nuance. Single-electrode recordings, sampling one neuron at a time, favored a serial feedforward view of vision. Later work showed the cortex to be densely recurrent, with feedback from higher areas shaping even V1 responses~\cite{hupe1998cortical} and recurrent processing necessary for computations a feedforward model cannot explain~\cite{felleman1991distributed,lamme2000distinct}. This finding did not make the feedforward account wrong, as rapid feedforward processing remains a real and central part of vision, but certainly enriched the knowledge.
The lesson for XAI is encouraging: imperfect tools can reveal a durable core of insight that better tools later refine rather than discard, so there is no need to wait for a perfect method before studying models.

\paragraph{\textbf{A converging model landscape.}} 
The space of computer vision models is shrinking in architectural diversity as it grows in raw number. Computer vision has evolved from a fragmented landscape of convolutional architectures~\cite{krizhevsky2009learning,simonyan2014vgg,szegedy2015,he2016deep} to a smaller number of dominant, general-purpose backbones and training paradigms: the Vision Transformer~\cite{dosovitskiy2021vit}, language-supervised contrastive models such as CLIP and its sigmoid-loss successors~\cite{radford2021learning,zhai2023sigmoid}, the self-supervised DINO family, now in its third generation~\cite{caron2021emerging,oquab2023dinov2,simeoni2025dinov3}, masked image modeling (MAE)~\cite{he2022masked}, emerging predictive world-model objectives such as I-JEPA~\cite{assran2023ijepa}, and their extension to video~\cite{tong2022videomae,wang2023videomaev2,bardes2024vjepa,assran2025vjepa2}. Although these methods differ in supervision, objectives and downstream capabilities, they increasingly share a common transformer backbone and produce broadly comparable visual representations. This convergence shifts the bottleneck for XAI. Instead of developing new explanation methods for ever-changing architectures, it is now feasible to use existing XAI tools to systematically compare how different training objectives and datasets shape internal representations and computations.

\paragraph{\textbf{Tools for comparing models.}}
The goal of model comparison is not to establish that models should converge on a single correct representation, but to characterize what they share, how they differ, and how those differences relate to properties we care about. Comparing models, however, is hard: each model's representation lives in its own idiosyncratic basis. Prior work has worked around this with aggregate comparisons, via the number of unique concept detectors~\cite{bau2017network,fong2018net2vec} or average human interpretability scores across a set of features~\cite{zimmermann2023,colin2024choosing}, or with soft matching of functionally analogous features~\cite{colin2026capability}. More recently, methods have emerged that support direct model comparison instead of aggregate or functional proxies, such as crosscoders and universal sparse autoencoders~\cite{lindsey2024crosscoders,gorton2024group,thasarathan2025universal,kassem2026delta}, which learn a concept space shared across models and allow comparison at a finer granularity. This converges with a fast-growing literature on measuring representational alignment across systems~\cite{kriegeskorte2008rsa,morcos2018cca,kornblith2019cka,williams2021generalized,sucholutsky2023getting,klabunde2025similarity}. Concept-based interpretability and representational alignment have largely developed as parallel tracks, but recent work shows they can be reconciled~\cite{dhimoila2026cosae}, making that literature particularly well suited to comparing models directly.

\section{Interpretability methods to study models}\label{sec:sparse}

A small but growing body of work already turns interpretability tools on models themselves. At the scale of one model, a detailed circuit analysis of InceptionV1's early layers~\cite{olah2020an,olah2020zoom,carter2019atlas} gave a qualitative account of what a vision network represents, surfacing features like curve detectors; an approach later turned on CLIP to reveal multimodal neurons that fire across both images and text~\cite{goh2021multimodal}; feature visualization scaled to modern architectures~\cite{fel2023unlocking} coupled with concept-based methods produced an explorable atlas of the concepts a ResNet50 uses across every class of ImageNet~\cite{imagenet_cvpr09}; and a 32,000-concept SAE dictionary for DINOv2~\cite{fel2026into} helped characterize the specific concepts recruited by different tasks: ``elsewhere'' concepts for classification, boundary concepts for segmentation, and three families of monocular depth cues for depth estimation that mirror classical visual neuroscience. The same lens has since been turned on the models now in widest use: {CLIP}'s image representation has been decomposed head by head and labeled in text~\cite{gandelsman2024clipdecomposition}, the artifact tokens of vision transformers reverse-engineered as internal computation registers~\cite{darcet2024registers}, and sparse autoencoders used to read interpretable, causally testable features out of both self-supervised backbones~\cite{stevens2025visionsae} and text-to-image diffusion models~\cite{surkov2024sdxlsae}.

\vspace{1mm}
Across models, the same tools address comparative questions. Interpretability dissociates from scale~\cite{zimmermann2023,zimmermann2024measuring} and, more generally, from capability~\cite{bau2017network,zimmermann2023,zimmermann2024measuring,colin2026capability}. It behaves as a stable, measurable property of a model, though how to steer it remains unclear~\cite{colin2026capability}. Direct representational comparison has its own long-standing toolkit (e.g., centered kernel alignment~\cite{kornblith2019cka} and model stitching~\cite{bansal2021stitching}) used, for instance, to ask whether vision transformers ``see'' like convolutional networks~\cite{raghu2021vitcnn}; more recent work mines the units common to an entire model zoo~\cite{dravid2023rosettaneurons} and labels neurons automatically to compare concept coverage across networks~\cite{oikarinen2023clipdissect}. Whether models converge on the same concepts gets a mixed answer~\cite{fong2018net2vec,mikriukov2023comparison,mustapha2024inter,dorszewski2025colors,thasarathan2025universal}, echoing a broader literature on representational convergence~\cite{huh2024platonic,groger2026aristotelian} and human--model alignment~\cite{muttenthaler2023humanalignment} that interpretability can both draw on and contribute to. Both outcomes are informative: convergence points to concepts that recur across models and may be especially fundamental, while divergence maps the variation across which interpretability itself may differ.

\vspace{1mm}
The human side of the question has its own, equally mature instrument: psychophysics. A growing body of work measures whether explanations improve human understanding of models, including forced-choice tasks that test whether people can identify which images a unit prefers~\cite{borowski2021,zimmermann2021},   studies examining whether interpretability methods help users predict model behavior or improve human--AI team performance~\cite{kim2021hive,colin2022cannot,nguyen2022visual,fel2023craft}, and broader evaluations of how well people grasp machine generated explanations~\cite{ribeiro2016lime,ghorbani2019towards,narayanan2018humans,lage2019evaluation,shen2020useful,nguyen2021effectiveness,sixt2022users,fel2023craft,casper2023red,rong2023towards}.
Across this literature, human understanding is shaped by the complexity and granularity of the information presented, and by how much information must be retained to form an adequate mental model~\cite{kulesza2013too,narayanan2018humans,lage2019evaluation,poursabzi2021manipulating,rong2023towards}.
To date, however, these methods have been used almost entirely to compare XAI methods.  Turning on models and asking which model a person can understand best is the natural, and still largely untaken, next step.

\vspace{1mm}
Part of the reason why this body of work remains limited may be structural. Machine learning venues tend to reward new methods and benchmark gains rather than careful, systematic evaluations of models. Tellingly, some of the most ambitious efforts to reverse-engineer what a network computes---the Circuits program and the work on monosemanticity---appeared as Distill articles and Transformer Circuits Thread posts rather than in peer-reviewed proceedings~\cite{olah2020zoom,cammarata2020thread,bricken2023monosemanticity,templeton2024anthropicv2}. This is beginning to change~\cite{fel2026into}, but the incentive to publish a systematic interpretability analysis of a model, or a head-to-head comparison across models, remains weak.

\section{A Research Agenda for Model-Centric XAI}\label{sec:agenda}

We close by laying out what taking this shift would require.

\paragraph{\textbf{First, agree on what understanding means.}} The field has never converged on a single definition of interpretability: the term is used to refer to different  properties~\cite{lipton2016mythos,doshivelez2017rigorous}, united primarily by the premise that it is ultimately human-centric~\cite{miller2017explanation}. This is not a semantic quibble: without agreement on what it means for a person to understand a model, there is no stable target for the comparisons, metrics, and proxies the rest of this agenda calls for. Establishing what constitutes understanding, and it should be demonstrated, is the prerequisite for everything that follows, and it is why automating its measurement before defining it (as discussed below) risks optimizing the wrong objective rather than a useful shortcut.

\paragraph{\textbf{Model-centric study is what drives method progress.}} Studying models directly is not a departure from developing new methods, it is a complementary way to identify what new developments are needed. A prime example of that is the deep dive into the neurons learned by InceptionV1~\cite{olah2020zoom}, which surfaced polysemantic units that resisted any single-neuron characterization, motivating the need for concept-based methods~\cite{elhage2022superposition,fel2023craft,gao2024scaling}. The same cross-validated study of a model that builds cumulative insight, as argued in Section~\ref{sec:now}, is also what exposes where our tools fall short. A model-centric agenda should treat this as a mechanism to use deliberately: sustained study of specific models is the most reliable way to discover what the field's toolbox is still missing.

\paragraph{\textbf{Where do models actually stand?}} This is the question interpretability was meant to answer, yet it remains largely unresolved. Despite a mature and increasingly sophisticated toolbox, relatively little work has applied these methods systematically across models (Section~\ref{sec:sparse}). As a result, we still do not know how much progress either approach---interpretable-by-design or post hoc interpretability---has made toward the original goal, nor which aspects of the problem remain unsolved. Answering this question is essential for directing future research. If the bottleneck lies in our interpretability methods, the priority should be developing better tools; if it lies in the models, the focus should shift toward designing and training models that are inherently more understandable. 

\paragraph{\textbf{Design and post hoc explanation remain uncompared.}} Interpretability-by-design and post hoc explanation have rarely been compared head to head, particularly with respect to the understanding they produce in human observers. By-design methods, ranging from B-cos networks~\cite{bohle2022bcos} to label-free concept bottlenecks~\cite{oikarinen2023labelfree}, explicitly impose structure meant to be interpretable, yet it remains largely unknown whether this structure leads to greater human understanding than post hoc explanations of unconstrained ``black-box'' models. Indeed, recent work suggests that the distinction between the two paradigms may be less fundamental than it appears: both can instantiate the same geometric object, differing only in how the object is identified~\cite{rocchi2025cones}. Whether either approach ultimately produces models that people understand better therefore remains an open empirical question which belongs squarely within a model-centric agenda.

\paragraph{\textbf{What predicts interpretability is still unresolved.}} Interpretability is dissociated from both scale and task performance~\cite{zimmermann2023,zimmermann2024measuring,colin2026capability}, which raises an obvious question: what properties predict it? Alignment with human perceptual or semantic organization is one candidate~\cite{fel2022harmonizing,sundaram2024dreamsimv2,muttenthaler2025nature,colin2026capability}, the locality of a feature's activations another~\cite{colin2026capability}, and robustness a third~\cite{ross2018improving,tsipras2018robustness,wang2022robust}. We are accumulating correlations about which models are more interpretable faster than we are developing principled accounts of why. If alignment does turn out to drive interpretability, it becomes more than a diagnostic: deliberately shaping model representations toward human ones---using perception, psychophysics, and neuroscience as reference points---could be a path to building interpretable models rather than only explaining them, closing the loop between measuring understanding and engineering it.

\paragraph{\textbf{Structural insight is the tractable place to start.}} Some of this agenda can be pursued without a human in the loop, making structural comparison the natural place to start. It is currently the most tractable direction and the best suited to broad, coordinated progress. Crosscoders and universal sparse autoencoders~\cite{lindsey2024crosscoders,thasarathan2025universal} now make it possible to compare representations and features across models directly, rather than relying on aggregate proxies or indirect functional matching~\cite{bau2017network,fong2018net2vec,colin2026capability}. The same tools also enable systematic comparisons between XAI methods that make different assumptions about how models organize  concepts~\cite{hindupur2025dual}, allowing disagreement between methods to become a source of insight rather than noise to be averaged away. More broadly, they provide a direct way to address the convergence question raised in Section~\ref{sec:sparse} directly: whether differently trained models converge on the same concepts is fundamentally a question of structural comparison, and the tools to answer it at scale are now within reach. 

\paragraph{\textbf{Human understanding remains the goal, and the bottleneck.}} Structural agreement between models is not the same as human understanding, and closing that gap is both the harder challenge and the ultimate objective of this agenda. Large-scale psychophysics has begun to shed light on what makes a representation understandable to human observers~\cite{zimmermann2023,zimmermann2024measuring, colin2026capability}, but scaling these evaluations remains prohibitively expensive. This limitation is becoming increasingly acute because tools such as crosscoders and universal sparse autoencoders can generate far more candidate representations than any human study can realistically evaluate. Developing scalable, validated proxies for human understanding, proxies trusted as substitutes for direct human evaluation without replacing it entirely, is a tractable and underexplored direction. It is also what would complete the arc of this paper: transforming the same interpretability toolbox that originally served only to explain models into an instrument for building models that people can understand.

\paragraph{\textbf{But it is too early to automate the human measure.}} The bottleneck above invites an obvious shortcut: replace the human with a model. The first serious attempt does exactly this: the Machine Interpretability Score automates the human forced-choice task~\cite{borowski2021} by scoring how consistently a learned perceptual-similarity model groups a unit's most activating images, and validates that score against human judgments before scaling it to millions of units~\cite{zimmermann2024measuring}. The scale is impressive, but the measure presupposes what it sets out to establish: the automated judge is itself a model of human perception, so interpretability collapses into agreement with a fixed proxy for what people find similar, which is precisely what we do not yet understand. The concern is not that such proxies lack value, but that they risk being adopted before the underlying construct is well defined. The field still has yet to answer more fundamental questions: what psychophysical task best captures understanding? What metric should determine whether a person truly understands a model? Encoding provisional answers into an automated evaluator risks prematurely fixing the definition of interpretability and ranking models by agreement with the proxy rather than by human understanding. The order matters: establish the human measure first, through principled psychophysics, and automate only once we know what we are automating. Even then, any automated proxy should remain anchored to, and regularly recalibrated against, direct human evaluation instead of replacing it.

\section{Conclusion}\label{sec:conclusion}

Explainable AI set out to make machine learning models understandable to humans, and for the past fifteen years it has largely focused on developing the tools to do so. We have argued that those tools are finally mature enough to return to the question that motivated them: not whether a given method explains a model, but whether our models are becoming interpretable and to whom. It is a subtle shift in perspective with far-reaching consequences. It recasts interpretability as a property of models, instead of methods, that can be measured, compared, and steered as models evolve.

The implications extend beyond interpretability research. As AI systems become more capable and broadly used, interpretability is fast becoming the foundation for trust, oversight, and certification. It is also our best opportunity to learn from these systems in return---to discover how they represent vision, language, and other domains where they now rival or surpass human performance. Systems neuroscience built a science around an object it could barely observe. We have the opposite predicament, models are fully accessible, yet they often remain opaque. Bridging the gap between complete access and genuine understanding is, we believe, the defining challenge of the next decade. The instruments are finally ready. What remains is the decision to turn them on the models and on the people they are ultimately meant to help.

\section*{Acknowledgment}
We are grateful to Thomas Fel, whose discussions and feedback over the years have helped shape many of the ideas developed here. T.S. acknowledges support from the Office of Naval Research (N00014-24-1-2026), the ONR REPRISM MURI (N00014-24-1-2603), the National Science Foundation (IIS-2402875), and the Artificial and Natural Intelligence Toulouse Institute (ANR-19-PI3A-0004). J.C. and N.O. acknowledge support from the Regional Government of Valencia, Spain (Conselleria de Industria, Turismo, Innovación y Comercio, Dirección General de Innovación), the European Union's Horizon Europe research and innovation program (ELIAS, grant agreement 101120237), and Intel Corporation; J.C. further acknowledges the Banco Sabadell Foundation.
\\ \\

\bibliographystyle{splncs04_unsrt}
\bibliography{main}
\end{document}